%% file: root.tex
\documentclass[letterpaper, 10 pt, conference]{ieeeconf}  

\IEEEoverridecommandlockouts                              

\usepackage[resetlabels]{multibib}
\usepackage{subfiles}
\usepackage{csquotes}
\usepackage{gensymb}
\usepackage{pifont}
\usepackage{textcomp}
\usepackage[table]{xcolor}
\usepackage{todonotes}
\usepackage{blindtext}
\usepackage{soul}
\usepackage{graphicx}
\usepackage{mathtools}
\usepackage{multirow}
\usepackage{flushend}
\usepackage{booktabs}
\usepackage{array}
\usepackage{tabularx}
\usepackage{threeparttable}
\usepackage{amsmath} 
\usepackage{amssymb} 
\usepackage{amsfonts}
\usepackage{bm} 
\usepackage{siunitx}
\usepackage{cancel}
\usepackage{microtype}
\usepackage{xcolor}
\usepackage{cite}
\usepackage[export]{adjustbox}
\usepackage[hang,flushmargin]{footmisc}
\usepackage{cuted}
\newcites{S}{References}
\usepackage[breaklinks,colorlinks]{hyperref}

\usepackage[capitalize]{cleveref}
\crefname{section}{Sec.}{Secs.}
\Crefname{section}{Section}{Sections}
\Crefname{table}{Table}{Tables}
\crefname{table}{Tab.}{Tabs.}

\definecolor{mygreen}{HTML}{00A64F}
\definecolor{myred}{HTML}{ED1B23}

\newcommand{\secref}[1]{Sec.~\ref{#1}}
\renewcommand{\eqref}[1]{Eq.~(\ref{#1})}
\newcommand{\figref}[1]{Fig.~\ref{#1}}
\newcommand{\tabref}[1]{Tab.~\ref{#1}}

\newcommand{\net}{BALViT}

\title{\LARGE \bf
Label-Efficient LiDAR Semantic Segmentation with\\ 2D-3D Vision Transformer Adapters
}

\author{Julia Hindel$^*$, Rohit Mohan$^*$, Jelena Bratuli\'c, Daniele Cattaneo, Thomas Brox, and Abhinav Valada
\thanks{* Authors contributed equally to this work.}%
\thanks{Department of Computer Science, University of Freiburg, Germany.}
\thanks{This work was funded by the Deutsche Forschungsgemeinschaft (DFG, German Research Foundation) – SFB 1597 – 499552394.}%
}

\begin{document}

\maketitle
\thispagestyle{empty}
\pagestyle{empty}

\begin{abstract}
LiDAR semantic segmentation models are typically trained from random initialization as universal pre-training is hindered by the lack of large, diverse datasets. Moreover, most point cloud segmentation architectures incorporate custom network layers, limiting the transferability of advances from vision-based architectures. Inspired by recent advances in universal foundation models, we propose \net, a novel approach that leverages frozen vision models as amodal feature encoders for learning strong LiDAR encoders. Specifically, \net~incorporates both range-view and bird's-eye-view LiDAR encoding mechanisms, which we combine through a novel 2D-3D adapter. While the range-view features are processed through a frozen image backbone, our bird's-eye-view branch enhances them through multiple cross-attention interactions. Thereby, we continuously improve the vision network with domain-dependent knowledge, resulting in a strong label-efficient LiDAR encoding mechanism.
Extensive evaluations of \net~on the SemanticKITTI and nuScenes benchmarks demonstrate that it outperforms state-of-the-art methods on small data regimes. We make the code and models publicly available at \mbox{\url{http://balvit.cs.uni-freiburg.de}}.
\end{abstract}


\input{sections/01_introduction}
\input{sections/02_related-work}

\input{sections/03_methodology}
\input{sections/04_experiments}
\input{sections/05_conclusion}


\footnotesize
\bibliographystyle{IEEEtran}
\bibliography{references}

\end{document}

%% file: sections/01_introduction.tex
\section{Introduction}
\label{sec:introduction}
\begin{figure}
    \centering
    \includegraphics[width=\linewidth]{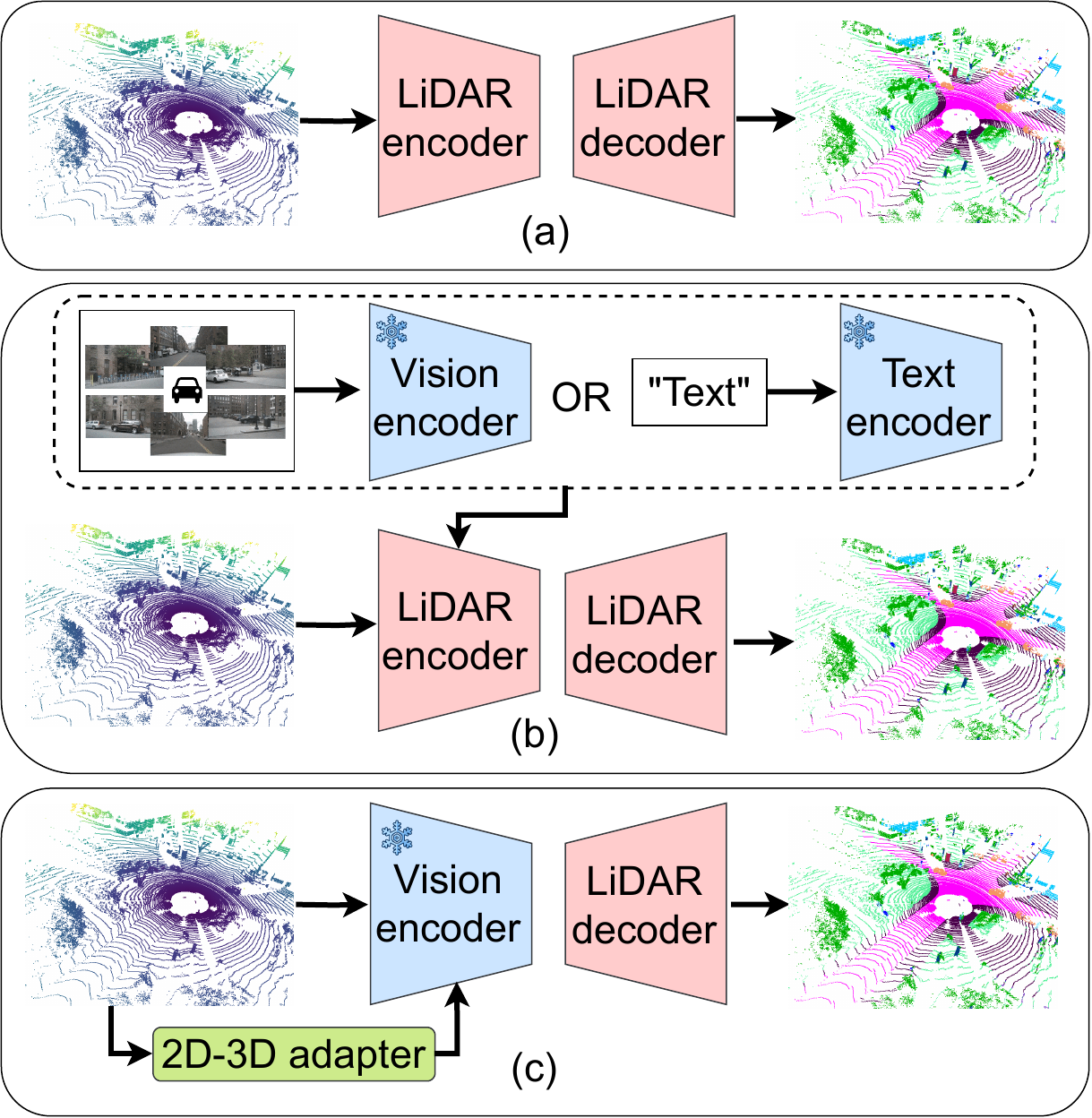}
    \caption{Different learning paradigms for LiDAR semantic segmentation. Learned modules are colored in red and frozen components in blue. (a)~Traditional methods leverage randomly initialized LiDAR networks. (b)~Vision or language foundation models are employed to distill knowledge into tailored LiDAR architectures. (c)~Transferring a pre-trained vision model into the LiDAR domain using a range-view projection and a 2D-3D adapter (ours).}
    \label{fig:teaser}
    \vspace{-0.4cm}
\end{figure}

Self-driving vehicles often rely on LiDAR sensors to semantically perceive their surroundings due to their invariance to lighting conditions~\cite{mohan2024progressive,lang2024point}.
However, LiDAR semantic segmentation methods typically follow a fully supervised learning paradigm~\cite{liu2023seal, nunes23tarl}, which requires labor-intensive data collection and manual labeling of large-scale point cloud datasets.
Recently, self-supervised methods have been introduced, enabling the pre-training of neural networks without the need for ground truth labels. Pre-training improves the performance of the network when fine-tuned on a small labeled dataset~\cite{nunes23tarl, nunes22, liu2023seal}. While these techniques excel with images~\cite{gosala2023skyeye,lang2024self,hindel2023inod} and natural language, the performance improvements are limited for LiDAR data. This is primarily due to the absence of large and diverse pre-training datasets that cover the significant domain shifts between different LiDAR sensors~\cite{bevsic2022unsupervised, xiao23survey}.
Consequently, label-efficient techniques that leverage models pre-trained on other modalities are required.
Prior work in this direction focuses on extending vision and language foundation models~\cite{oquab2024dino2, radford2021clip} for 3D perception by adding a separate network branch~\cite{chen23clip2scene,zhou23uni3d,hegde23,zeng23, zhu23pointclipv2} or directly distilling network features~\cite{liu2023seal,xu2024superflow}, as shown in~\figref{fig:teaser}. However, these methods still train a randomly initialized LiDAR sub-network and require time-synchronized camera and LiDAR streams with accurate extrinsic calibration. 

Another line of work follows the paradigm of a universal foundation model where only the patch embedding and the decoder are tailored to 3D point clouds~\cite{ando23rangevit, liang24pointgst, zhou24dynamic}. Consequently, these methods can benefit from strong pre-trained feature extractors and can easily leverage advances in the field of foundation models~\cite{ando23rangevit}. However, RangeViT~\cite{ando23rangevit} shows the need for full fine-tuning of vision models for outdoor LiDAR segmentation, which is computationally expensive and infeasible when only a small number of labeled scans are available. Further, we demonstrate that adapters and visual prompt tuning methods initially introduced for 2D computer vision tasks do not readily generalize to label-efficient outdoor 3D point cloud segmentation. Consequently, we highlight the need for domain-aware adapters that incorporate the geometric constraints of 3D sparse point clouds to transform a pre-trained vision model into a robust 3D segmenter.

In this paper, we introduce \net~(\textbf{B}ird-Eye-View \textbf{A}dapted \textbf{L}iDAR \textbf{ViT}), which seamlessly incorporates vision models with a novel 2D-3D adapter for label-efficient LiDAR semantic segmentation.
In particular, we extend the vision model-based network RangeViT~\cite{ando23rangevit} with a lightweight, polar Bird's-Eye View (BEV)-based adapter. We reason that vision models excel at detecting object shapes, whereas our adapter incorporates geometric reasoning. Finally, we introduce a separate BEV decoder branch to address and correct misclassifications, ensuring the model generates accurate 3D semantic segmentation by refining the output through complementary branches.

Our main contributions can be summarized as follows:
\begin{itemize}
    \item \net, a novel LiDAR semantic segmentation architecture that leverages pre-trained vision models as a backbone.
    \item A newly-designed 2D-3D adapter for label-efficient refinement of vision models for sparse 3D segmentation.
    \item Extensive evaluations and ablation study on two datasets and tasks with a focus on minimal labeled data.
    \item Publicly available code and pre-trained models at \url{http://balvit.cs.uni-freiburg.de}.
\end{itemize}

%% file: sections/02_related-work.tex
\section{Related Work}\label{sec:related-work}
{\parskip=0pt
\noindent\textit{LiDAR Semantic Scene Segmentation}: 
Point cloud semantic segmentation networks are commonly categorized according to the chosen representation. Voxel-based architectures sub-sample points according to coordinate grids before applying 3D convolutions~\cite{zhu2021cylindrical} or radial attention~\cite{lai2023spherical}. On the other hand, point-based methods extract features directly from points using multi-layer perceptrons~\cite{qi16pn} or transformer-based architectures~\cite{ptv3, ptv2}. However, both voxel-based and point-based approaches are computationally expensive and leverage custom architectures that do not benefit from advances in 2D computer vision research. The third category of LiDAR segmentation models convert point clouds into 2D projections, enabling the integration of standard computer vision architectures~\cite{zhang2020polarnet, ando23rangevit, Xu2023FRNetFN}. PolarNet proposes to map point cloud features into a BEV projection before applying circular padded convolutions~\cite{zhang2020polarnet}. RangeVit~\cite{ando23rangevit}, FRNet~\cite{Xu2023FRNetFN}, and RangeFormer~\cite{kong23rangeformer} convert the point cloud into multichannel range projections. Further, RangeViT~\cite{ando23rangevit} motivates the development of LiDAR architectures that closely resemble 2D vision architectures and pairs LiDAR-specific patch embeddings and decoder layers with a standard ViT backbone for semantic segmentation. However, RangeViT requires end-to-end fine-tuning on 3D data and suffers from the perspective distortion of range-views. Consequently, we propose to extend RangeViT with a parameter-efficient 2D-3D adapter to enhance the geometric prediction capabilities, effectively improving performance in small data settings.}

{\parskip=2pt
\noindent\textit{LiDAR Representation Learning}: 
Self-supervised representation learning focuses on pre-training a LiDAR segmentation model with unlabeled data. While SegContrast~\cite{nunes22} and TARL~\cite{nunes23tarl} perform contrastive learning between class-agnostic segments, BEVConstrast~\cite{bevcontrast} applies its contrastive loss between BEV projections of two augmented point clouds. Masked autoencoders, on the other hand, are trained for surface reconstruction~\cite{zhang23iscc} or occupancy map prediction~\cite{boulch23also}. 
Another family of approaches leverages geometric and time-aligned camera-LiDAR sensor input in addition to vision foundation models to pre-train 3D segmentation models. Commonly, the vision model extracts class-agnostic segments and supervises the 3D branch with their respective cross-modal alignment in feature space~\cite{puy2024pillars, sautier22slidr, xu2024superflow, liu2023seal}. Another direction of research extends the vision-language model CLIP~\cite{radford2021clip} with a separate LiDAR branch, which is trained with multi-modal contrastive learning~\cite{chen23clip2scene, hegde23, xue23ulip}. However, the outlined approaches require aligned multi-modal sensor input and only focus on training a LiDAR encoder from scratch. Additionally, self-supervised representation learning requires a large amount of data, which is not readily available in the LiDAR domain, where large domain gaps exist between various sensors. We argue that leveraging pre-trained vision transformers directly is more efficient and suitable for small data problems.}

{\parskip=2pt
\noindent\textit{Parameter-Efficient Fine-tuning}: 
This research field aims to adapt strong foundation models to a specific task with minimal parameter updates. A common practice is to restrict weight updates to bias terms only~\cite{jia22vpt}. Alternatively, visual prompt tuning (VPT)~\cite{jia22vpt} inserts additional parameters in the form of learnable prompts into the transformer backbone. The family of LoRA adaptation~\cite{hu22lora} focuses on continuously attaching trainable rank decomposition matrices in parallel to a frozen transformer block. To enhance performance on dense prediction tasks, ViT-Adapter~\cite{chen23va} introduces a spatial prior module as a separate convolutional stem that interacts with the main network branch through cross-attention. 

For dense 3D point clouds, various adapters target the transformation of 3D pre-trained models (e.g., Point-BERT~\cite{yu21pointbert}) for single-object classification and part segmentation. IDPT~\cite{zha23idpt}, PPT~\cite{fei2024pcpt}, and DAPT~\cite{zhou24dynamic} enhance pre-trained models with learnable prompts.
PointGST~\cite{liang24pointgst} motivates the usage of a spectral input representation to enhance the spatial correlation of 3D pre-trained tokens~\cite{liang24pointgst}.
However, prior work shows that frozen vision or language foundation backbones paired with 3D-tailored adaptation mechanisms can surpass self-supervised 3D pre-training for single-object 3D segmentation~\cite{tang25anypoint, li24af, hegde23, huangepcl24}. For example, Any2Point projects a point cloud into multiple 2D views and introduces an adapter tailored for local spatial aggregation and weighted feature ensembling~\cite{tang25anypoint}. On the other hand, PointFormer shows the benefit of applying a standard 2D adaptation mechanism~\cite{li24af}.
However, existing adapter-based methods are limited to single-object, dense point cloud settings. Motivated by this line of work, we propose \net{} which is, to the best of our knowledge, the first 2D-3D adapter for sparse, outdoor point cloud processing. Our adapter enhances a frozen vision model with spatial priors retrieved from two complementary point cloud representations.}

%% file: sections/03_methodology.tex
\begin{figure*}
    \vspace{0.1em}
    \centering
    \includegraphics[width=\linewidth]{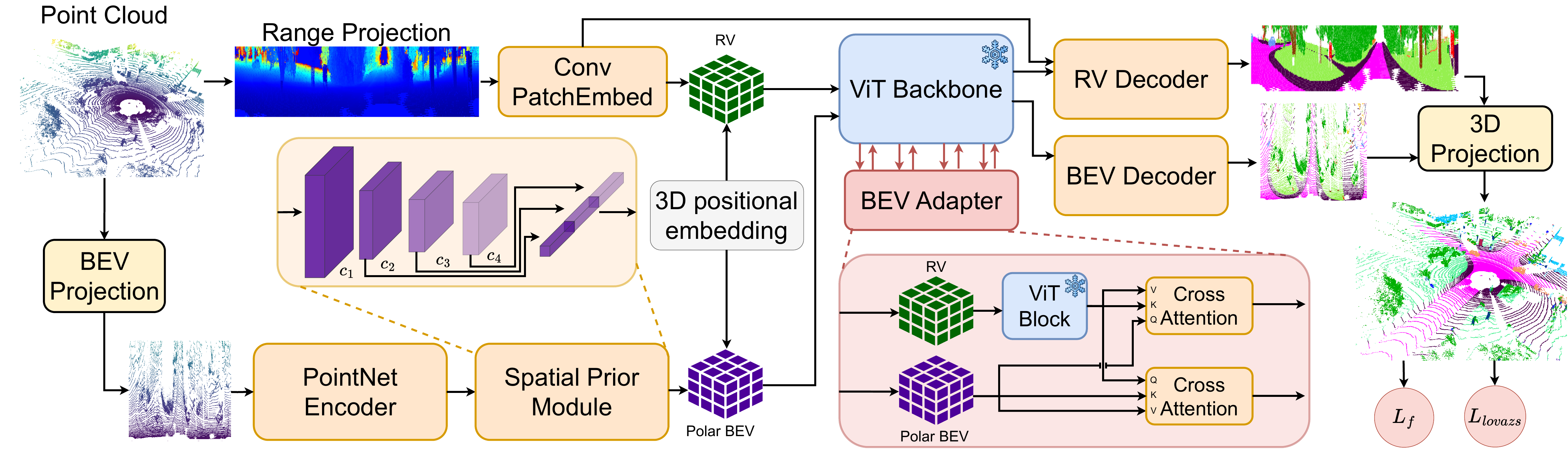}
    \caption{Our network \net~encodes a point cloud in orthogonal range-view (RV) and bird-eye-view (BEV) branches. Our spatial prior module converts the BEV branch into multi-scale features, which interact with the RV branch during its traversal of the frozen ViT backbone. Last, our two decoders independently obtain pointwise class labels from the respective feature maps. \looseness=-1}
    \label{fig:model}
\end{figure*}

\section{Technical Approach} \label{sec:technical-approach}
Our proposed \net\ is tailored for adapting pre-trained vision transformer (ViT) backbones for LiDAR semantic segmentation. We argue that these architectures enable amodal feature encoding, which we enhance with our label-efficient 2D-3D adapter. Specifically, we first encode the point cloud in two separate branches, a range-view (RV) and a BEV encoder.
Next, the RV features are processed by a frozen ViT backbone. During this backbone traversal, our novel 2D-3D adapter enables bidirectional feature enhancement between RV and BEV representations. Finally, we decode each feature branch with separate 3D decoders to combine the strengths of both views, effectively reducing misclassifications. \figref{fig:model} provides an overview of our architecture. In \secref{sec:rv_emb} and \secref{sec:bev_emb}, we outline the employed point cloud representations and encoding branches. Then, we introduce our proposed 2D-3D adapter in \secref{sec:ad}, and finally we provide a description of the decoders in \secref{sec:dec}.

\subsection{Range-View Encoding}\label{sec:rv_emb}
The input to our architecture is a LiDAR point cloud $P \in \mathbb{R}^{N \times 4}$ composed of $N$ points, each with four values $(x, y, z, i)$. The variables $(x, y, z)$ refer to the Cartesian coordinates and $i$ represents the observed intensity of the returned LiDAR beam.
To convert the LiDAR point cloud into a range projection of size $H \times W$, we first compute the pixel coordinates for each point $p_j \in P$ as follows:
\begin{equation}
\label{f_rangeview}
\begin{bmatrix}
h_j \\[0.1cm] w_j
\end{bmatrix}
=
\begin{bmatrix}

\frac{1}{2} \left(1 - \arctan(y_j, x_j) \frac{1}{\pi} \right) W \\[0.1cm]
\left( 1 - \left( \arcsin(z_j, \frac{1}{r_j}) - \frac{f_{\text{down}}}{f_v} \right) \right) H
\end{bmatrix} ,
\end{equation}
where $r_j = \sqrt{x_j^2 + y_j^2 + z_j^2}$ corresponds to the range of the point $p_j$ and $f_{v} = f_{\text{up}} - f_{\text{down}}$ being the vertical view of the LiDAR sensor~\cite{ando23rangevit}. We then construct the RV image $I^{RV}$ according to \cref{eq_rv} below:
\begin{equation} \label{eq_rv}
    I^{RV}_{h_j,w_j} = \left[ r_j, z_j, i_j \right] .
\end{equation}

When multiple points are mapped to the same pixel within $I^{RV}$, we select the one with the smallest range.
We posit that range-views are best suited for direct processing with a pre-trained ViT backbone since this representation most closely resembles camera images. To encode images, a ViT backbone~\cite{dosovitskiy20vit} commonly applies a single linear projection. To bridge the domain gap between camera images and $I^{RV}$, we replace this layer with our Conv PatchEmbed block that is composed of four residual blocks inspired by SalsaNext~\cite{cortinhal20salsanext}. We later refer to features after this operation as $RV_{stem}$. We employ average pooling with a kernel size of ($PH_{RV}$,$PW_{RV}$) and a final $1 \times 1$ convolution to convert $RV_{stem}$ to have $D_{RV}$ channels. Finally, we add a classification token and 2D positional embedding $E_{2D} \in \mathbb{R}^{(M+1) \times D_{RV}}$ to transfer the flattened $M$ features into a suitable format for further encoding by the ViT backbone~\cite{ando23rangevit}. 

\subsection{Polar Bird-Eye View Encoding}\label{sec:bev_emb}
We complement the RV features with an orthogonal BEV branch. Specifically, we use a polar BEV representation $I^{BEV}$ with grid cell dimension $H_{BEV}$, $W_{BEV}$, and $Z_{BEV}$. Therefore, we first convert the ($x_j,y_j,z_j$) coordinates of each point $p_j$ into polar coordinates according to:
\begin{equation}
\label{f_polar}
\begin{bmatrix} \rho_j \\ \phi_j \\ \theta_j \end{bmatrix} = 
\begin{bmatrix} \sqrt{x_j^2 + y_j^2} \\ -\arctan2(y_j, -x_j) \\ \frac{(z_j-z_{min})}{z_{max} - z_{min}} Z_{BEV}\end{bmatrix}
\end{equation}
where $z_{min}$ and $z_{max}$ represent the minimum and maximum height, respectively.
Then, we randomly select $N_p$ points per BEV grid cell and leverage a PointNet-inspired encoder~\cite{qi16pn} consisting of 4 blocks of fully-connected layers, BatchNorm, and ReLU to compute an embedding of dimension $D_{BEV}$ for each point. Next, we perform max pooling over the \( Z_{BEV} \)-dimension to obtain our BEV features of dimensionality ($D_{BEV}$, $H_{BEV}$, $W_{BEV}$). We create multi-scale BEV features with our spatial prior module to improve the model's capability to capture contextual information. This module consists of a ResNet-inspired stem and three convolutional blocks to transform the acquired BEV features into multi-scale representations $c1$, $c2$, $c3$, $c4$ with resolutions ($H_{BEV}/2$, $W_{BEV}/2$),($H_{BEV}/4$, $W_{BEV}/2$), ($H_{BEV}/8$, $W_{BEV}/2$) and ($H_{BEV}/8$, $W_{BEV}/2$). Further, our spatial prior module transforms the channel dimension to $D_{RV}$ in order to align with the RV features. 

Subsequently, we add a 2D learnable positional embedding to the multi-scale features to preserve information about their respective scales when flattening them for further interactions in the 2D-3D adapter. We argue that these orthogonal LiDAR encoding mechanisms, RV-based and BEV-based, are crucial for our method since they learn complementary point cloud representations whose combination transforms the ViT backbone into a strong LiDAR encoder.

\subsection{2D-3D Adapter Module}\label{sec:ad}
Before combining the two encoder branches, we add 3D positional embeddings $E_{3D} \in \mathbb{R}^{(M+1) \times D_{RV}}$ to both feature maps. Our sinusoidal positional embedding is computed separately in $x$, $y$, and $z$ dimensions following the original transformer's positional encoding~\cite{vaswani2017attention}.
Then, we concatenate the component-wise embeddings and upscale the resulting vector with two $1 \times 1$ convolutions to obtain a positional embedding of dimension $D_{RV}$.
We leverage the transformations in \eqref{f_rangeview} and \eqref{f_polar} to extract the Cartesian coordinates of each feature in the two views, RV and BEV. We then use these geometric positions to infer their positional embeddings and add them to the respective feature maps independently.
Our 3D positional embedding ensures that the interactions between the feature maps take into account the spatial geometries of the 3D scene. We emphasize the need for this positional embedding in~\secref{ab}.

The RV features are further encoded through subsequent blocks of the frozen ViT backbone. We perform feature integration of our two branches (RV and BEV) at different layers of the ViT backbone using our tailored interaction modules. Each interaction entails two parallel injector modules where one injector module (INJ) updates the features of one network branch (X) by performing sparse cross-attention with the other branch (Y) following~\cref{eq:crossatn}, where $\gamma_i$ is a learnable parameter. The second injector performs the same operation but with reversed branches.
\begin{equation}\label{eq:crossatn}
X_i = X_i + \gamma_i \text{Attention}(\text{norm}(X_i), \text{norm}(Y_i))
\end{equation}

Our parallel stacked cross-attentions enhance the RV features from the frozen ViT backbone branch with domain-specific knowledge from the BEV branch (X=RV, Y=BEV), while at the same time, the BEV features are refined with novel context from the RV features (X=BEV, Y=RV). We note that it's crucial that the interactions are performed in parallel to maximize the information flow between the two feature branches. Specifically, this operation ensures that information is continuously encoded and refined without one branch being more prominent. We ablate different configurations of the cross-attention layer in \secref{ab}. We repeat these interactions after layers 5, 11, 17, and 23 of the ViT backbone. Since the backbone is frozen, only the weights of the interaction modules are updated, which results in our method being parameter-efficient. Consequently, we emphasize the suitability of our method for small datasets. 

\subsection{Decoder}\label{sec:dec}
We attach separate LiDAR decoders to the RV and BEV network branches to predict semantic segmentation on each view independently.
For the RV branch, we use the decoder proposed by RangeViT~\cite{ando23rangevit}, which is composed of a convolutional decoder, a PixelShuffle layer, and a 3D refiner.
For the BEV branch, we propose a progressive decoder to combine the multi-scale BEV feature maps.
Our decoder fuses multi-scale features in a hierarchical manner: \( c_4 \) is first combined with \( c_3 \), then upsampled and merged with \( c_2 \), and finally upsampled again before integrating with \( c_1 \).
Last, we upsample the feature map to the original BEV input resolution using PixelShuffle~\cite{shi2016real}, which helps maintain the integrity of point clouds.
We then use three blocks of convolutions to recover the $Z_{BEV}$-dimension and predict a semantic class for every BEV cell grid. Finally, we employ a grid sampler to reproject our polar BEV predictions from polar ($S$, $W_{BEV}$, $H_{BEV}$, $Z_{BEV}$) to 3D coordinates ($x$, $y$, $z$) as detailed in \secref{sec:bev_emb}, where $S$ is the number of classes.

We train our network using multi-class Focal~\cite{Lin2017FocalLF} and Lov{\'a}sz-Softmax~\cite{berman2018lovasz} losses separately for each decoder branch's pointwise predictions.
During inference, we merge pointwise RV and BEV predictions according to a predefined threshold $t$:
\begin{equation}
\text{Output} =
\begin{cases}
\hat{y}_{\text{RV}}, & \text{if } \hat{y}_{\text{RV}} > t \\
\hat{y}_{\text{RV}} & \text{if } \hat{y}_{\text{RV}} \leq t \text{ and } \hat{y}_{\text{BEV}} < t \\
\hat{y}_{\text{BEV}} , & \text{if } \hat{y}_{\text{RV}} \leq t \text{ and } \hat{y}_{\text{BEV}}  > t \\
\end{cases},
\end{equation}
where \( \hat{y}_{\text{RV}} \) and \( \hat{y}_{\text{BEV}} \) are the highest scores of the RV and BEV predictions, respectively.
With the introduced merging mechanism, we can correct misclassifications in the RV feature branch, which results in a further boost in performance as presented in \secref{ab}.

%% file: sections/04_experiments.tex
\section{Experimental Evaluation}
In this section, we quantitatively and qualitatively evaluate
the performance of \net{} on LiDAR semantic segmentation and
also provide a comprehensive ablation study to demonstrate the significance of our contribution. We first present an overview of the used datasets and then describe the experimental settings.

\subsection{Datasets}
We evaluated our proposed method on two autonomous driving datasets, SemantiKITTI and nuScenes.

{\parskip=2pt
\noindent\textit{SemanticKITTI~\cite{skitti}}: This dataset is derived from the KITTI Odometry benchmark and consists of a total of 19,130 training and 4,071 validation scans. The dataset was collected with a single Velodyne HDL-64E LiDAR, which is mounted on a car driving around the city of Karlsruhe, Germany. The dataset provides point-wise labels for 19 semantic classes.}

{\parskip=2pt
\noindent\textit{nuScenes~\cite{nuscenes}}: This dataset consists of 28,130 training and
6,019 validation scans, which have been recorded in 1,000 scenes of 20 seconds each in Boston and Singapore. The LiDAR sensor that was used has 32 laser beams, and 16 semantic classes are annotated.}

\begin{table*}[!t]
\centering
\caption{Label-efficient training results on 0.1\%, 1\%, 10\% of the SemanticKITTI and nuScenes datasets.}
\label{tab:main}
\vspace*{-0.2cm}
\setlength{\tabcolsep}{0.3cm}
\begin{threeparttable}
\begin{tabular}{llp{0.1cm}cccp{0.1cm}ccc}
\toprule
& & & \multicolumn{3}{c}{\textbf{SemanticKITTI mIoU (\%)}} & & \multicolumn{3}{c}{\textbf{nuScenes mIoU (\%)}}  \\ \cmidrule{4-6} \cmidrule{8-10}
& \multirow{-2}{*}{Method} & & 0.1\% & 1 \%   & 10\%  & & 0.1\% & 1 \%   & 10\% \\ 
\midrule
\multirow{4}{*}{\rotatebox{90}{\shortstack{Fully- \\ Supervised}}} & Minkowski SR-Unet18~\cite{Choy194DSC} & & - & 39.50 & - && - & 30.30 & 56.15 \\
& FRNet~\cite{Xu2023FRNetFN} && 30.09 & 40.78 & 61.55  && 28.03 & 48.98 & 69.99 \\
& SphereFormer~\cite{lai2023spherical} && 29.21 & 42.81 & 58.81 && 30.42 & 50.06 & 69.25 \\
& RangeViT~\cite{ando23rangevit}  && 28.74 & 43.53 & 58.53 && 27.79 & 52.88 & \textbf{71.84} \\
\midrule
\multirow{4}{*}{\rotatebox{90}{\shortstack{Vision \\ Distillation}}} & SLidR~\cite{sautier22slidr} && - & 44.60 & - && - & 38.30 & 59.84 \\
& ST-SLidR~\cite{Mahmoud23stslidr} && - & 44.72 & - && - & 40.75 & 60.75 \\
& SEAL~\cite{liu2023seal} && - & 46.63 & - && - & 45.84 & 62.97 \\
& CLIP2Scene~\cite{chen23clip2scene} && - & 42.60 & - && - & 56.30 & - \\
\midrule

\multirow{6}{*}{\rotatebox{90}{\shortstack{Parameter- \\ Efficient \\ Fine-Tuning}}} & Frozen ViT backbone         && 29.97 & 45.91 & 58.10  && 28.72 & 54.70  & 63.28\\
& Bias tuning         && 30.86 & 45.63 & 58.14 && 28.15 & 56.05 & 65.08  \\
& LoRA~\cite{hu22lora}              && 31.65 & 46.27 & 59.53  && 28.27 & 57.57 & 66.38  \\
& VPT~\cite{jia22vpt} && 31.07 & 46.08 & 58.38 && 29.68 & 55.67 & 65.52 \\
& Vit Adapter~\cite{chen23va}   && 29.55 & 45.01 & 57.43  && 27.50 & 56.06 & 67.71 \\
& \net\ (Ours)  && \textbf{32.85} & \textbf{51.80} & \textbf{61.91}  && \textbf{31.86} & \textbf{59.27} & 70.13 \\
\bottomrule
\end{tabular}
The results are reported on the val set, and all metrics are in [\%].
\end{threeparttable}
\vspace{-.3cm}
\end{table*}

\subsection{Experimental Setup}\label{setup}
We use RV projections of size ($64, 2048$) with the patch size being defined as $PW_{rv}=8$ and $PH_{rv}=2$. We set the dimensionality of our RV features to $D_{rv} = 384$. Our polar BEV representation contains $480 \times 360 \times 32$ grid cells for the full point cloud, with each feature having $D_{bev}=512$ dimensions. We define $z_{min}$=-1 and $z_{max}$=3. We use a Cityscapes pre-trained ViT-S backbone for our model and show the effect of different pretrainings in \secref{ab:backbone}. We train our networks and baselines for $100$, $500$, and $1000$ epochs with dataset splits of $0.1$\%, $1$\%, and $10$\%, respectively. We use the SLidR~\cite{Mahmoud23stslidr} protocol to create training subsets via uniform sampling, e.g. selecting every 100th frame for the 1\% subset. We use a batch size of $4$, a learning rate of $0.0004$, and a cosine learning rate scheduler with $20$ warm-up epochs. Further, we take random RV crops of size ($64$, $384$) for SemanticKITTI and ($64$, $768$) for nuScenes. Consequently, we equally subsample our BEV features, resulting in polar grid sizes of 
$480 \times 68  \times 32$ and $ 480 \times 135 \times 32$, respectively. We augment the point clouds with random horizontal flips with a probability of $0.5$ and scale the point cloud in ranges of [$0.95$-$1.05$]. Further, we randomly resample rare class points. For nuScenes, we additionally randomly rotate in all three axes in the range of $-5$ and $5$ degrees with a 50\% probability. We set the inference threshold $t$ to $0.9$.

\subsection{Quantitative Results}
We compare \net{} with four fully supervised and four vision model distillation approaches. We select fully supervised methods with model size comparable to our method and which achieve top performance on the SemanticKITTI 100\% setting. We use the respective author’s published code and apply the same augmentations outlined in \secref{setup}. The self-supervised methods SLidR~\cite{sautier22slidr}, ST-SLidR~\cite{Mahmoud23stslidr}, SEAL~\cite{liu2023seal} and CLIP2Scene~\cite{chen23clip2scene} pretrain the Minkowski SR-Unet18 architecture~\cite{Choy194DSC} on the entire nuScenes dataset. For these methods, we report the achieved performance from their respective published work. We also compare with five parameter-efficient fine-tuning methods that we integrated in the RangeViT architecture~\cite{ando23rangevit}. For a fair comparison, we apply the same training configurations as used in our method.

All models are evaluated with the mean intersection-over-union (mIoU) metric. Results on SemanticKITTI and nuScenes are presented in \tabref{tab:main}. Notably, \net{} outperforms all supervised and self-supervised baselines by at least 1.44pp on the 0.1\% and 1\% settings. This can be attributed to the strong vision priors from the pretrained ViT, which our model effectively enhances via our novel 2D-3D adapter.
However, the difference in performance decreases for the 10\% settings, indicating that LiDAR-specific transformers possess strong encoding capabilities but require a sufficiently large quantity of labeled data.
We further analyze the model performances when training with higher quantities of data in \secref{sec:100d}.
In the last block of rows of \tabref{tab:main}, we compare several parameter-efficient methods applied to the same base architecture as used in our method. First, we find that parameter-efficient fine-tuning is preferred to full-finetuning (RangeViT) in all the evaluated scenarios. We conclude that 2D pre-trained networks provide strong priors for 3D segmentation, which benefits from enhancement rather than retraining. Our method outperforms other parameter-efficient fine-tuning methods by at least 1.80pp on the 1\% and 10\% training settings. We find that our novel method of combining RV and BEV features within a frozen ViT model adds domain-specific knowledge in the feature encoding process, which outperforms standard parameter-efficient vision approaches. We discuss the effects of different components of our method in~\secref{ab:add}. 

\subsection{Ablation Study}\label{ab}
In this section, we study the impact of various network components and dataset settings on the performance of our approach. We perform all ablation experiments on the 1\% SemanticKITTI setting unless otherwise stated. 

\subsubsection{Influence of Network Components}\label{ab:add}
We sequentially analyze the effect of our proposed network components in \tabref{tab:ab_elem}. In the first row, we display the achieved performance when using the RangeViT architecture with a frozen ViT backbone. Next, we show that applying our 2D-3D adapter with unaligned single-scale features results in a performance improvement of $2.62$pp, which we attribute to the enhancement of RV features through our BEV feature branch. Further, the performance is improved by an additional $0.53$pp when RV and BEV features are aligned with our introduced 3D positional embedding strategy. We reason this observation with an improved cross-attention between geometric-aligned features. Next, we emphasize the importance of creating multi-scale BEV features and show that the observed performance is further improved by $0.63$pp. Finally, we observe an improvement of $2.1$pp when incorporating our BEV decoder into the network. We find that miss projections and semantic miss-classification can be reduced when guiding the prediction by two independent decoders.

\begin{table}
\centering
\caption{Ablation study on various components of \net{} on 1\% SemanticKITTI training setting.}
\vspace*{-0.2cm}
\label{tab:ab_elem}
\begin{threeparttable}
\begin{tabular}{ccccp{0.05cm}c}
 \toprule
2D-3D & & & BEV & &\\
Adapter & \multirow{-2}{*}{3D PE} & \multirow{-2}{*}{SPM} & decoder && 1\%\\
\midrule
& & & && 45.91 \\
\checkmark & & & && 48.53 \\
\checkmark & \checkmark & & && 49.06 \\
\checkmark & \checkmark & \checkmark &  && 49.70 \\
\checkmark & \checkmark & \checkmark & \checkmark && \textbf{51.80} \\
\hline
\end{tabular}
  \begin{tablenotes}[para,flushleft]
       \footnotesize
       3D PE: 3D positional embedding, SPM: Spatial Prior Module.
  \end{tablenotes}
\end{threeparttable}
\end{table}

\subsubsection{Influence of Pre-trained Vision Backbone}\label{ab:backbone}
We compare the performances when using different backbone pretrainings in \tabref{tab:skitti-bb}. In line with \cite{ando23rangevit}, we find that using a Cityscapes pretrained network results in the highest score. We posit that \net{} benefits from leveraging a vision backbone pre-trained on a closely related training domain (i.e., visual autonomous driving). Further, we record the lowest performance when leveraging a random initialized but trainable ViT backbone, which underscores the essential role of pretrained vision models when only a small amount of labels (1\%) is available. 
We observe the second lowest performance on a Depth Anything pre-trained model, which we attribute to the difference in target task, depth estimation instead of semantic segmentation. 

\begin{table}[!t]
\centering
\caption{Comparison with different vision backbones on 1\% SemanticKITTI training setting.}
\vspace*{-0.2cm}
\begin{threeparttable}
\begin{tabular}{lp{0.05cm}c}
 \toprule
Backbone initialization && 1 \% \\ 
\midrule
Supervised random init.$\dagger$   && 44.58  \\
Depth Anything~\cite{depthanything24} && 45.11 \\
Dino~\cite{caron21dino}   && 46.94 \\
Dino~v2~\cite{oquab2024dino2}   && 47.65 \\
MoCov3~\cite{chen2021mocov3}  && 48.47 \\
Supervised ImageNet~\cite{deng2009imagenet} &&  48.68 \\
Cityscapes~\cite{Cordts2016Cityscapes}  && \textbf{51.80}\\
\bottomrule
\end{tabular}
\begin{tablenotes}[para,flushleft]
       \footnotesize
       $\dagger$: vision backbone is updated during training.
     \end{tablenotes}
\end{threeparttable}
\label{tab:skitti-bb}
\vspace{-.3cm}
\end{table}

\subsubsection{Influence of Injector Modules}
As described in \secref{sec:ad}, we perform our two injector operations (INJ) in parallel to enhance the information flow between the two feature branches. In this section, we study the order of operations in our 2D-3D adapter module. Specifically, we analyze the impact of applying one injector before a ViT block and the second INJ after the same block in \tabref{tab:skitti-decoder}. 
In the first row, we refine the BEV features before the block and the RV feature after, while in the second row we reverse the order.
In the third row, we show that applying both injection operations in parallel results in the highest performance. These results show that parallel injectors enhance the learning of different aspects simultaneously, which allows the network to capture more diverse features.

\begin{table}[!t]
\centering
\caption{Ablation study on varying sequences of injector (INJ) on 1\% SemanticKITTI training setting. }
\vspace*{-0.2cm}
\begin{tabular}{lp{0.05cm}c}
 \toprule
Injector strategy && 1 \% \\ 
\midrule
INJ$_{bev,rv} \rightarrow$ ViTBlock $\rightarrow$ INJ$_{rv,bev}$ &&   46.17 \\
INJ$_{rv, bev} \rightarrow$ ViTBlock $\rightarrow$ INJ$_{bev,rv}$ &&  48.42 \\
ViTBlock $\rightarrow$ INJ$_{rv, bev} \| $INJ$_{bev,rv}$ &&  \textbf{51.80} \\
\bottomrule
\end{tabular}
\label{tab:skitti-decoder}
\end{table}

\subsubsection{Influence of Increased Training Data}\label{sec:100d}
We emphasize that our method is designed for small data regimes, whereas tailored LiDAR architectures benefit from training on more data, as shown in \tabref{tab:sk100}. Thus, we argue that when large quantities of data are available, LiDAR-specific models are preferred. Nevertheless, we show that our method also outperforms full fine-tuning of the RangeViT architecture when training on 100\% of the dataset, confirming the impact of our 2D-3D adapter and independent decoder branches. 

\begin{table}[!t]
\centering
\caption{Performance on 100\% SemanticKITTI training data.}
\vspace*{-0.2cm}
\begin{threeparttable}
\begin{tabular}{lp{0.05cm}c}
\toprule
Method && 100 \% \\ 
\midrule
SphereFormer&& 67.8$\dagger$ \\
FRNet && \textbf{68.7}$\dagger$ \\
\midrule
RangeViT~\cite{ando23rangevit}    && 60.28 \\
Frozen ViT          && 60.51 \\
Bias tuning~\cite{jia22vpt} && 61.94 \\
LoRA~\cite{hu22lora}  && 59.53 \\
VPT~\cite{jia22vpt} && 61.46 \\
Vit Adapter~\cite{chen23va} && 61.29 \\
\net{} (Ours) && 62.37\\
\bottomrule
\end{tabular}
\begin{tablenotes}[para,flushleft]
       \footnotesize
       $\dagger$: performance recorded in published work.
     \end{tablenotes}
\end{threeparttable}
\label{tab:sk100}
\vspace{-.3cm}
\end{table}

\begin{figure*}
  \centering
  \footnotesize
  \setlength{\tabcolsep}{0.05cm}
      {\renewcommand{\arraystretch}{1}
        \begin{tabular}{p{0.5cm}p{2.5cm}p{2.5cm}p{2.5cm}p{0.5cm}p{2.5cm}p{2.5cm}p{2.5cm}}
        \multicolumn{4}{c}{SemanticKITTI} & \multicolumn{4}{c}{nuScenes} \\ \cmidrule{2-4} \cmidrule{6-8}
        & \multicolumn{1}{c}{Ground Truth} & \multicolumn{1}{c}{LoRA} & \multicolumn{1}{c}{\net~(ours)} & & \multicolumn{1}{c}{Ground Truth} & \multicolumn{1}{c}{LoRA} & \multicolumn{1}{c}{\net~(ours)} \\

        a) & \includegraphics[width=\linewidth]{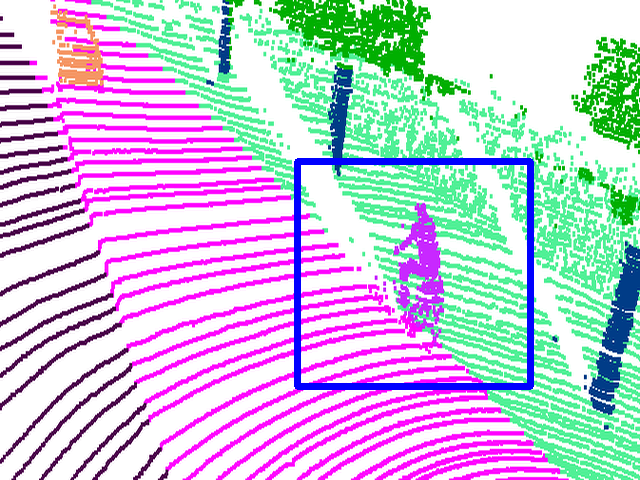} & 
      \includegraphics[width=\linewidth]{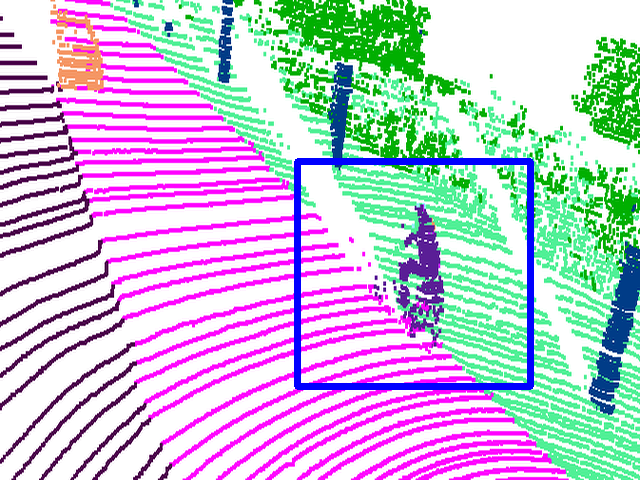} &
        \includegraphics[width=\linewidth]{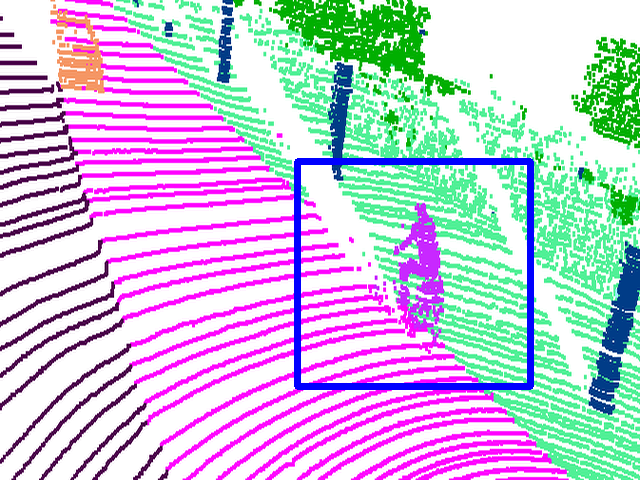} & e) &
      \includegraphics[width=\linewidth]{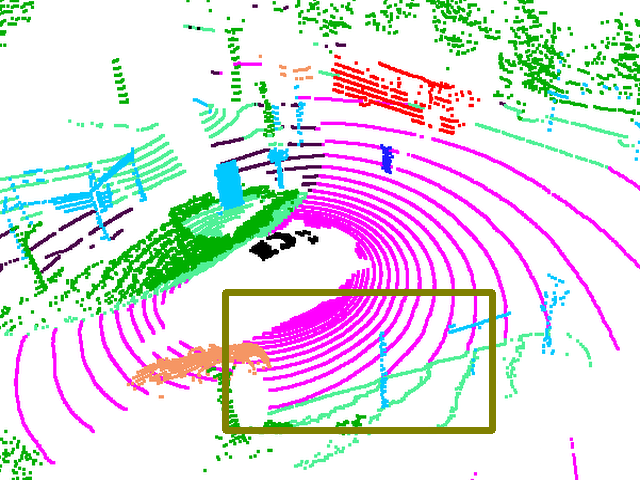} &  
       \includegraphics[width=\linewidth]{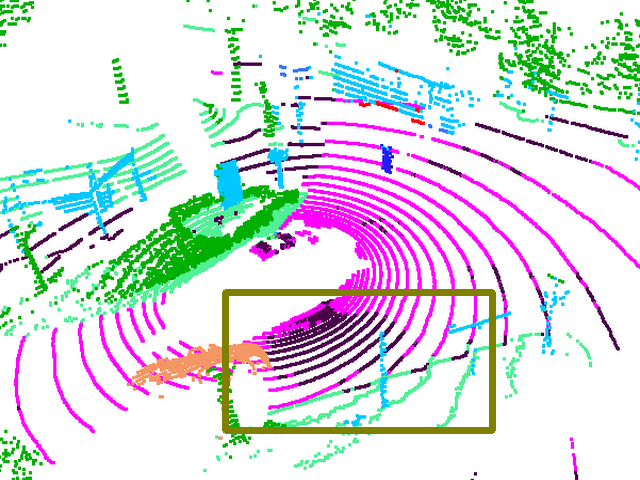} &
      \includegraphics[width=\linewidth]{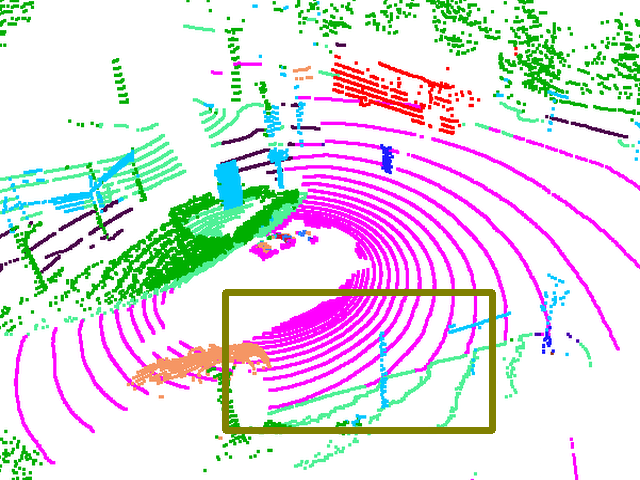} \\

        b) &    \includegraphics[width=\linewidth]{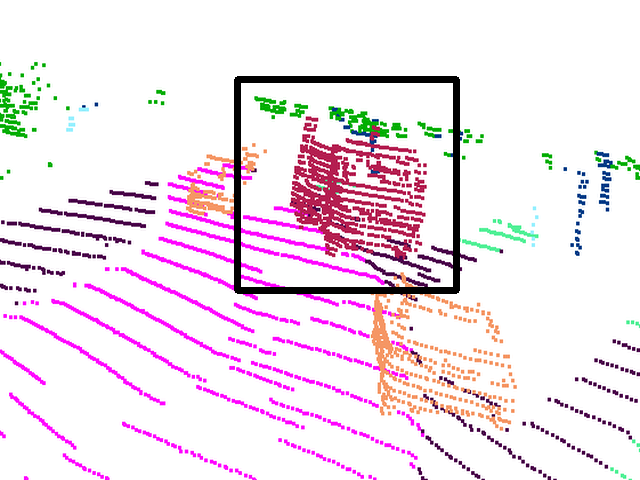} & 
      \includegraphics[width=\linewidth]{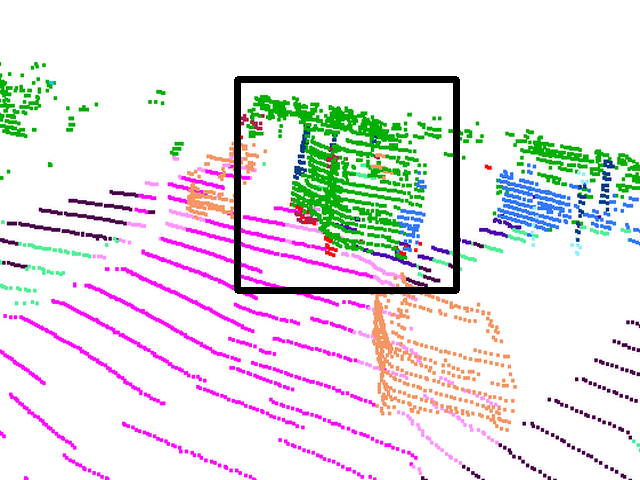} &
        \includegraphics[width=\linewidth]{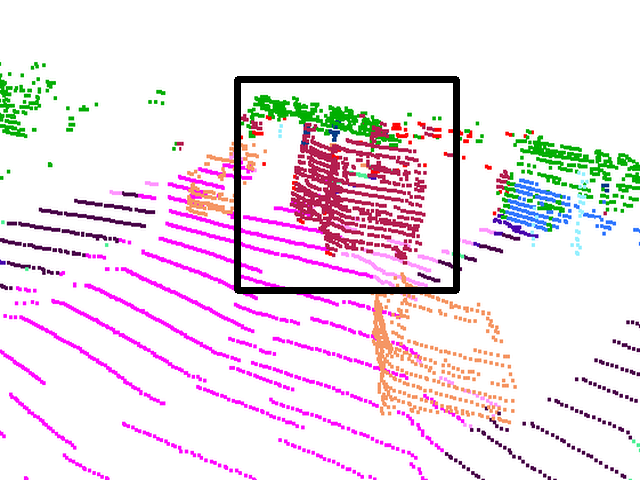} & f) &
      \includegraphics[width=\linewidth]{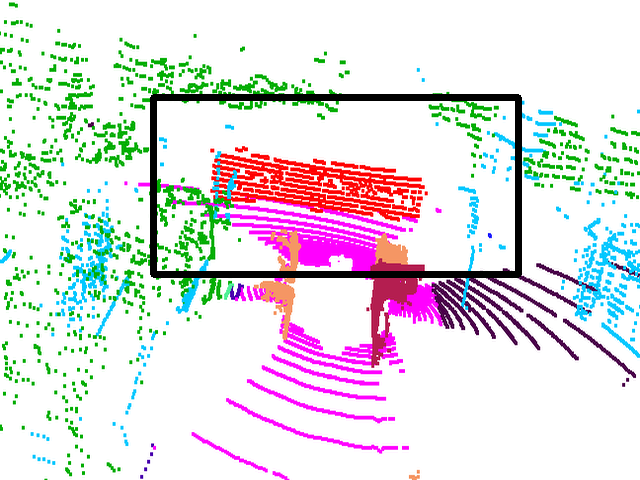} &  
       \includegraphics[width=\linewidth]{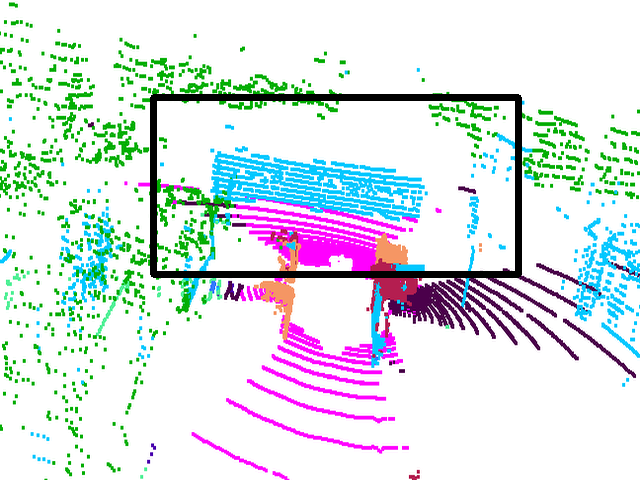} &
      \includegraphics[width=\linewidth]{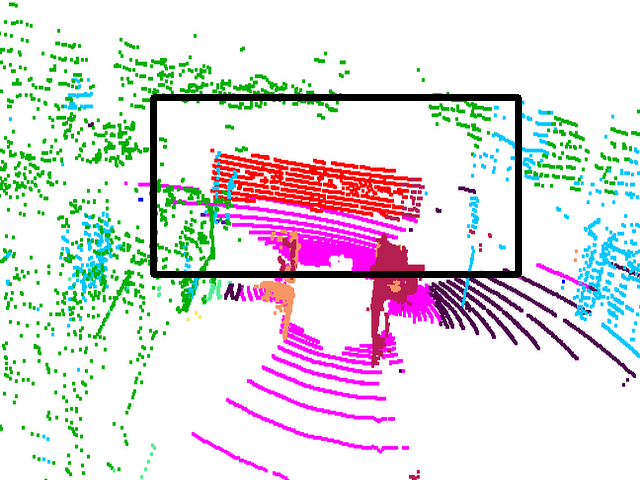} \\

      c) & \includegraphics[width=\linewidth]{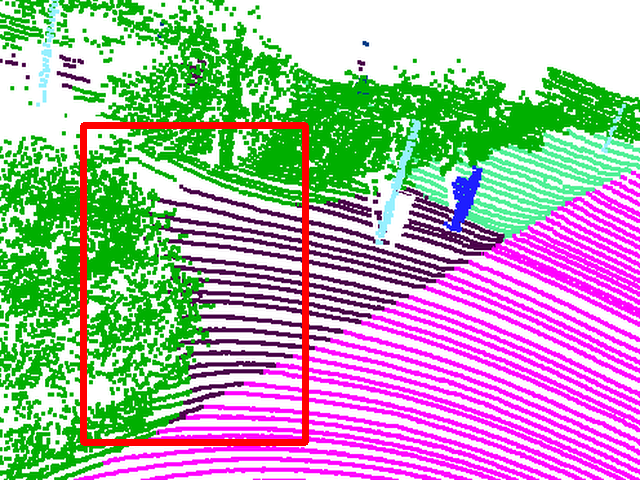} & 
      \includegraphics[width=\linewidth]{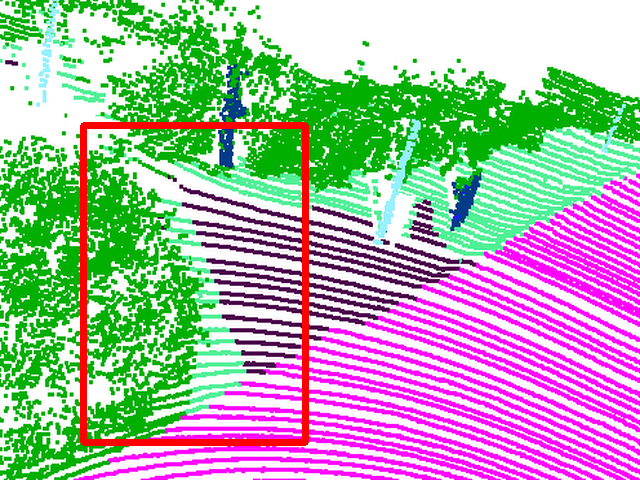} &
       \includegraphics[width=\linewidth]{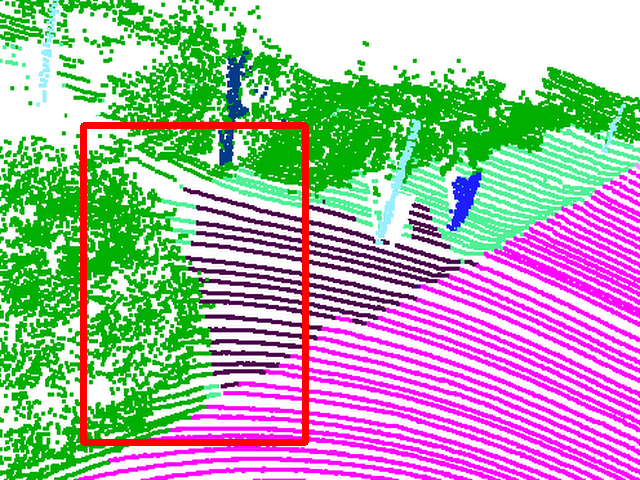}& g) &
      \includegraphics[width=\linewidth]{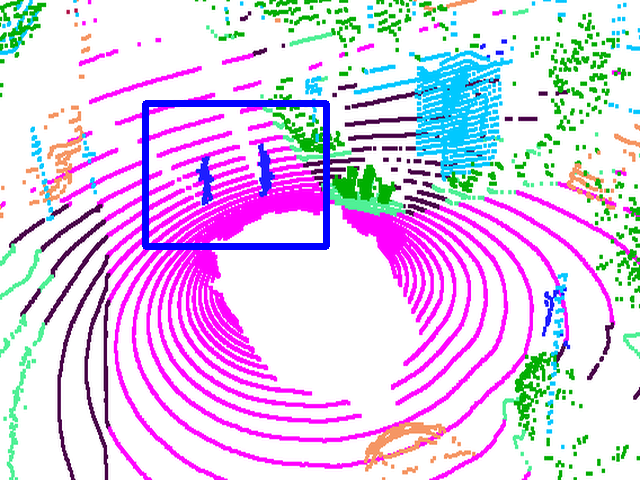} &  
       \includegraphics[width=\linewidth]{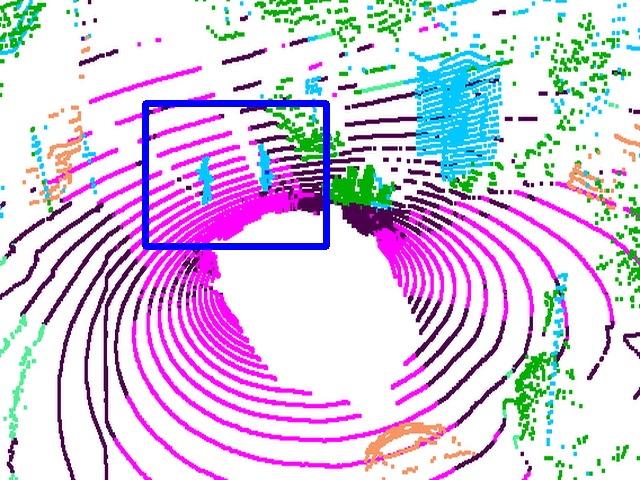} &
      \includegraphics[width=\linewidth]{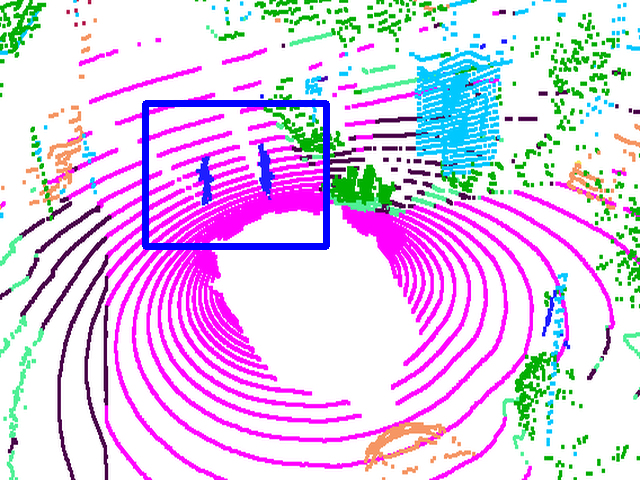} \\

    d) & \includegraphics[width=\linewidth]{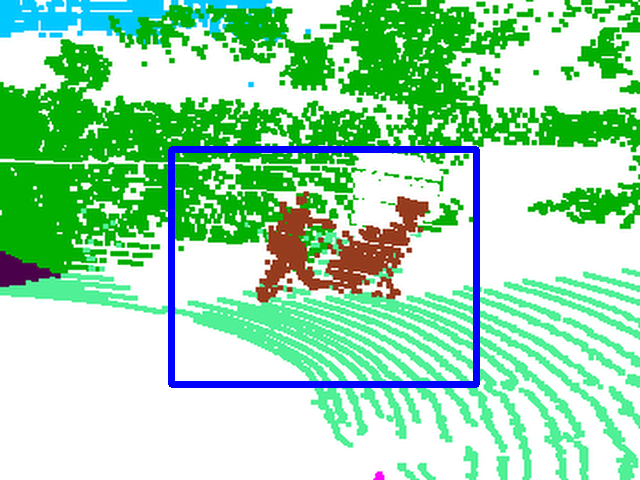} & 
      \includegraphics[width=\linewidth]{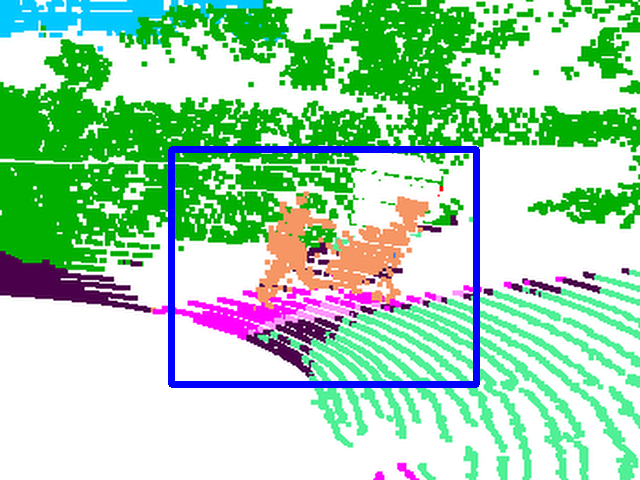} &
        \includegraphics[width=\linewidth]{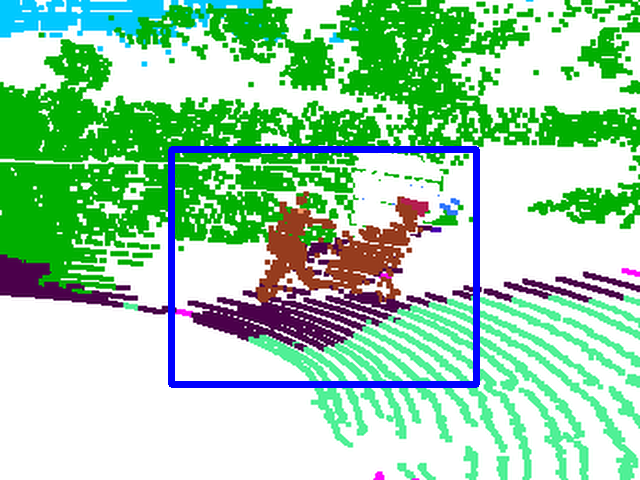} & h) &
      \includegraphics[width=\linewidth]{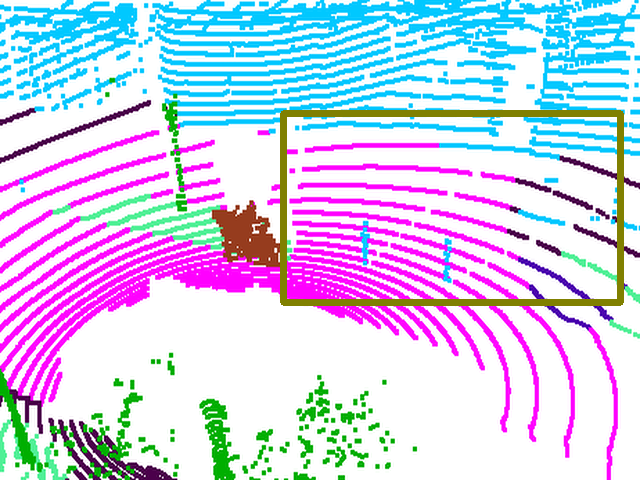} & 
       \includegraphics[width=\linewidth]{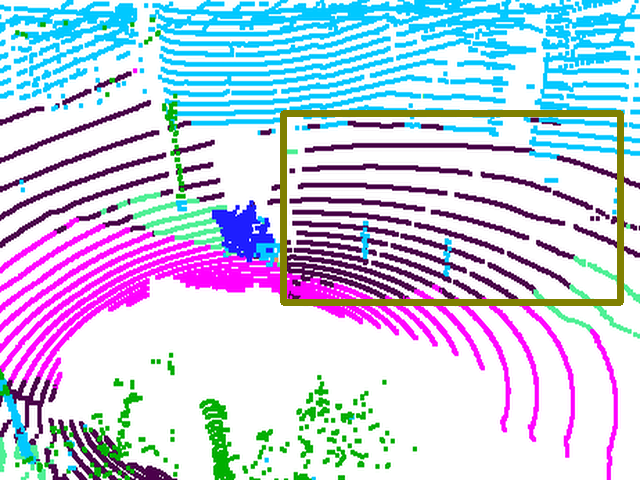} &
      \includegraphics[width=\linewidth]{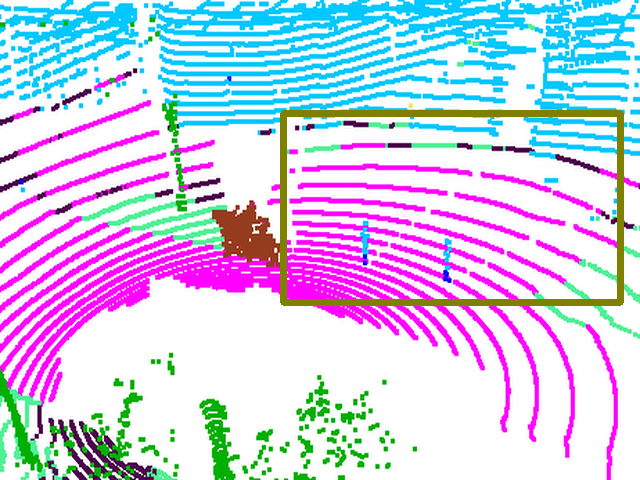} \\
    \end{tabular}}
      \caption{Qualitative results of \net~on LiDAR semantic segmentation on nuScenes and SemanticKITTI.}
      \label{fig:qualitative}
      \vspace{0.3cm}
\end{figure*}

\subsection{Qualitative Results}

We further evaluate the impact of \net{} by qualitatively comparing its predictions with the second best-performing parameter-efficient method LoRA~\cite{hu22lora} in \figref{fig:qualitative}. Our qualitative analysis confirms the improved performance of \net{}. On both datasets, our method yields more coherent segmentation, resulting in stable prediction, less ``patchified'' regions, and fewer misclassifications. 
While LoRA misclassifies entire objects as marked blue in a), d), and g), \net{} effectively predicts the correct label for the complete instance due to our independent decoder branches, which counteract misclassifications of rare classes. Further, our method better captures finer semantic differences between terrain and vegetation as marked in red in c). However, we note that both LoRA and our method misclassify vegetation as trunk in c), which demonstrates that both methods are constrained by the diversity of input samples in this label-efficient setting. 

We also observe that \net{} shows better performance on segmenting larger objects (e.g., truck, bus and other-vehicles) as shown in black in b), and f), which likely arises from an improved localization guidance. We find that our method improves semantic knowledge in label-efficient training regimes by aligning BEV and RV representations. Next, we observe much stronger localization and boundary-aware segments on the nuScenes dataset in e) and h) (green markings), where class boundaries for the semantic categories of sidewalk and road are more refined. We attribute this improvement to our multi-scale BEV features, which assist in capturing fine details.

%% file: sections/05_conclusion.tex
\section{Conclusion}\label{sec:conclusion}
We present \net, a novel method for LiDAR semantic segmentation tailored for small data regimes. Our approach encodes a RV projection with a frozen ViT backbone which we enhance with our 2D-3D adapter. Subsequently, we merge the predictions of our separate RV and BEV decoders for improved performance on rare classes. We observe that our approach outperforms existing state-of-the-art supervised and self-supervised baselines on label-efficient training settings of 0.1\% and 1\% on the SemanticKITTI and nuScenes datasets. The proposed method is one of the early works to show that 2D vision models provide valuable priors for LiDAR semantic scene segmentation. Consequently, we motivate future work to directly integrate pre-trained vision architectures in point cloud segmentation models to leverage future advancements in foundation model research.